\documentclass[10pt,twocolumn,letterpaper]{article}

\usepackage{wacv}
\usepackage{times}
\usepackage{epsfig}
\usepackage{graphicx}
\usepackage{amsmath}
\usepackage{amssymb}

\wacvfinalcopy % *** Uncomment this line for the final submission

\def\wacvPaperID{232} % *** Enter the wacv Paper ID here

\newcommand\blfootnote[1]{%
  \begingroup
  \renewcommand\thefootnote{}\footnote{#1}%
  \addtocounter{footnote}{-1}%
  \endgroup
}

\ifwacvfinal\fi
\begin{document}

%%%%%%%%% TITLE
\title{Global-Local Airborne Mapping (GLAM): \\
Reconstructing a City from Aerial Videos}

% Authors at the same institution
%\author{First Author \hspace{2cm} Second Author \\
% Authors at different institutions
\author{Hasnain Vohra \\
Vision Systems Inc.\\
{\tt\small hasnain.vohra@visionsystemsinc.com}
\and
Maxim Bazik \\
Vision Systems Inc.\\
{\tt\small max.bazik@visionsystemsinc.com}
\and
Matthew Antone \\
Sciopic Inc.\\
{\tt\small antone@sciopic.com}
\and
Joseph Mundy \\
Vision Systems Inc.\\
{\tt\small jlm@visionsystemsinc.com}
\and
William Stephenson \\
Massachuessets Institute of
Technology\textsuperscript{*}\\
{\tt\small wtstephe@mit.edu}
}

\maketitle
\ifwacvfinal\thispagestyle{empty}\fi

\blfootnote{\textsuperscript{*}Work performed while at Vision Systems Inc.}

%%%%%%%%% ABSTRACT
\begin{abstract}
Monocular visual SLAM has become an attractive practical approach for robot localization and 3D environment mapping, since cameras are small, lightweight, inexpensive, and produce high-rate, high-resolution data streams. Although numerous robust tools have been developed, most existing systems are designed to operate in terrestrial environments and at relatively small scale (a few thousand frames) due to constraints on computation and storage.

In this paper, we present a feature-based visual SLAM system for aerial video whose simple design permits near real-time operation, and whose scalability permits large-area mapping using tens of thousands of frames, all on a single conventional computer. Our approach consists of two parallel threads: the first incrementally creates small locally consistent {\it submaps} and estimates camera poses at video rate; the second aligns these submaps with one another to produce a single globally consistent map via factor graph optimization over both poses and landmarks. Scale drift is minimized through the use of 7-degree-of-freedom similarity transformations during submap alignment.

We quantify our system's performance on both simulated and real data sets, and demonstrate city-scale map reconstruction accurate to within 2 meters using nearly $90,000$ aerial video frames - to our knowledge, the largest and fastest such reconstruction to date.
\end{abstract}

%%%%%%%%% BODY TEXT
\section{Introduction}
\label{sec:intro}

Recent progress in unmanned aerial vehicle (UAV) technology has led to exponential growth in the use of drones with onboard sensing. Originally designed for military applications, camera-equipped UAVs are now commonplace in domains such as commercial surveillance, photography, disaster management, product delivery, and mapping \cite{colomina2014unmanned}. An essential step toward better utilization of aerial videos and autonomous drones is real-time localization of the vehicle and accurate mapping of the observed world. Localization relying solely on onboard inertial and GPS measurements, however, cannot achieve pixel-level accuracy due to error accumulation, latency, and relatively coarse precision; furthermore, onboard sensing can be unreliable in GPS-denied environments. Visual simultaneous localization and mapping (SLAM) attempts to address these challenges using camera images to augment or replace other sensors.

Our proposed method, nicknamed \textit{Global-Local Airborne Mapping} (GLAM), approaches large-scale visual SLAM by partitioning the camera's video stream into small local \textit{submaps} created using point features, epipolar geometry between keyframes, point triangulation, and incremental bundle adjustment (BA). Submaps are aligned globally using a graph-based least squares optimization that minimizes the distance between corresponding 3D points. Small temporal overlap ensures that sequential submaps observe common scene content and hence provides correspondences for alignment, while restricting submap size to a small number of frames mitigates drift accumulation. Associations between non-overlapping submaps, for instance due to loop closures, are detected via fast bag-of-visual-words recognition. This paper demonstrates that such a system can run in near real-time on videos of the scale of a hundred thousand frames.

\section{Related Work}
Monocular SLAM is a long-researched subject that has seen evolution from filter based \cite{thrun2005probabilistic, davison2007monoslam} to keyframe based \cite{strasdat2012visual, resch2015scalablesfm} approaches. 
PTAM \cite{klein07parallel}, originally developed for augmented reality applications, was one of the first widely used and practical real-time SLAM systems, but was limitated in terms of scale and robustness.  Subsequent research improved upon this point feature and bundle adjustment methods \cite{strasdat2010scale, resch2015scalablesfm}. More sophisticated monocular SLAM approaches like DTAM \cite{newcombe2011dtam} and LSD-SLAM \cite{engel12iros} perform semi-dense reconstruction by optimizing directly on image intensities rather than discrete features.  SVO \cite{Forster2014ICRA} is a semi-direct approach monocular SLAM approach designed specifically for Micro Aerial Vehicles (MAVs); although SVO operates at very high frame rates, it does not perform loop closure and has been evaluated only on small datasets. ORB-SLAM \cite{mur2015orb} is a recent feature-based approach that has shown good performance on a wide variety of datasets, but largely untested for large aerial datasets. 

City-scale 3D reconstruction has received some attention in the structure-from-motion research community \cite{musialski2013survey}. Pollefeys \textit{et al.} \cite{pollefeys2008detailed} reconstruct parts of a city from a hundred thousand frames using INS and metadata to simplify the computation. Agarwal \textit{et al.} \cite{agarwal2011building} and Heinly \textit{et al.} \cite{heinly2015_reconstructing_the_world} perform city and world scale reconstruction respectively from large collections of photographs utilizing cloud computing. Google Earth provides 3D models of a few selected cities, but the technology behind its reconstructions is largely unknown. Leotta \textit{et al.}\cite{leotta2016open} have published open-source software for aerial SfM focused on videos of a scale of a thousand frames but do not address scaling beyond that.

Real-time reconstruction of a city from aerial videos using purely visual data has remained largely unexplored in the literature: traditional visual SLAM approaches do not address the challenges of very long image sequences and large-scale maps, while structure-from-motion methods require substantial offline computation and/or information from additional sensors. Our system incorporates elements of both visual SLAM and large-scale 3D reconstruction, demonstrating fast and accurate city-scale mapping on aerial videos nearly 90,000 frames in length on a single consumer grade computer.

\begin{figure*}
\begin{center}
\includegraphics[width=1.0\linewidth]{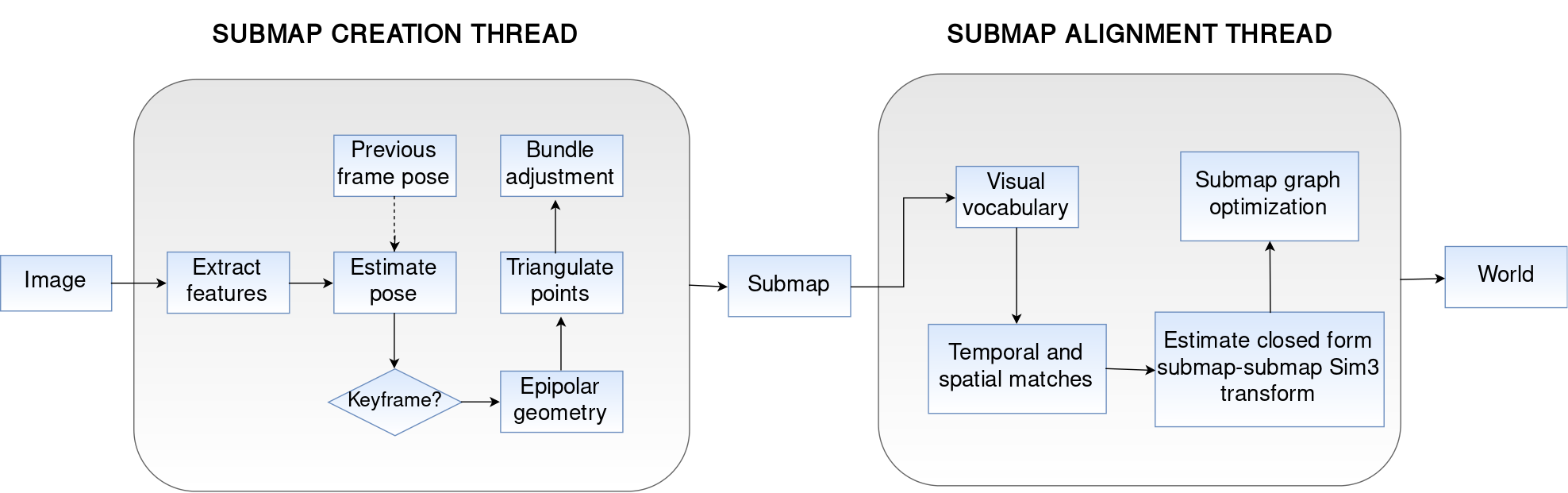}
\end{center}
   \caption{\textbf{GLAM} system overview, showing the main steps performed by submap creation and submap alignment threads.}
\label{fig:overview} 
\end{figure*}

\section{Proposed Approach}

Like many existing real-time SLAM systems, we adopt a multi-threaded strategy for computational efficiency, partitioning the problem into tasks that run in parallel. However, rather than defining the tasks as \textit{tracking} and \textit{mapping} \cite{klein07parallel, Forster2014ICRA, mur2015orb}, we define tasks as \textit{local submap creation} and \textit{global submap alignment}. Figure \ref{fig:overview} shows an overview of GLAM's building blocks.
%Such a division allows an incremental reconstruction of the observed world in two different threads. The runtime of the submap creation thread remains mostly constant and the submap alignment thread remains linear for a large number of frames 

The submap creation thread operates similarly to previous point features and keyframe based approaches \cite{klein07parallel, mur2015orb, strasdat2010scale} that extract point features from video frames, determine relative pose, triangulate points, and perform incremental bundle adjustment. Building the Hessian matrix for incremental bundle adjustment is the most computationally expensive step \cite{sibley2009adaptive}: as the number of frames processed increases, the cost of building the Hessian at every BA iteration becomes prohibitive for real-time operation. We therefore limit each submap to a small number of keyframes in order to maintain a relatively consistent and bounded processing rate.

%The submap creation thread implements a keyframe-based sparse 3D reconstruction using bundle adjustment. This thread extracts point features from video frames, matches them and determines relative pose of keyframes using the essential matrix. Points are triangulated using matches from the keyframes. After point triangulation, the cameras from the keyframes and the 3D points are bundle adjusted. The subsequent frames are then localized using the existing map until it becomes difficult to localize the frames. A keyframe is then identified, points are triangulated again and added to the map, followed by bundle adjustment. 
%TODO: FIX THIS PARAGRAPH

Each submap consists of a set of 3D landmarks. The submap alignment thread creates and optimizes a pose-graph to determine the set of 7-DoF similarity transformations, one per submap, that minimizes the distances between corresponding landmarks. In contrast to existing approaches that close loops by creating a global database of keyframes (e.g.\, \cite{mur2015orb}), our approach closes loops by building a visual vocabulary of submaps.

GLAM has been evaluated on both synthetic and real datasets. The former are created by simulating drone flight trajectories over an aerial LIDAR point cloud \cite{rigis}, while the latter consists of a large continuous aerial video (labeled ``Downtown'') with nearly $90,000$ 1-megapixel frames captured at $30$ fps.

The main contribution of this paper is a novel visual SLAM pipeline that (1) partitions work into parallel threads of fast local submap creation and large-scale global submap alignment; (2) has the ability to close large loops (3) reconstructs accurate city-size maps from aerial videos; and (4) operates in near real-time at the scale of a hundred thousand images.

\subsection{Local Submap Creation}

The following sections describe keyframe-based submap creation in greater detail. The central idea of keyframe-based SLAM is to use only frames that have sufficiently distinct information for 3D reconstruction. Each video frame is processed in sequence, with keyframes created periodically to reduce redundancy and improve efficiency---only the keyframes are used in intra-submap bundle adjustment.

\textbf{Feature Extraction.} Point features are extracted from each video frame. We use SIFT features \cite{lowe2004distinctive} throughout our system due to their proven robustness to viewpoint, scale and orientation changes. SIFT extraction and matching can be slow, so we use a GPU implementation \cite{wu2007siftgpu} for greater efficiency.

%Feature matching between two frames is done using a brute-force approach.
%The Euclidean distance between all features from both images are computed and are a two-%way match table is computed. 
%Features that have minimum distance between two images and have a distance more than a %threshold are used as a match.
%Feature matching is done by computing the Euclidean distance between the descriptors  and a normalized cross-correlation matrix of Euclidean distances is computed.

\textbf{Tracking.}
Existing 3D landmarks are projected into the current frame using the pose of the previous frame, since inter-frame motion is assumed to be small. Points that fall outside the frame or that were originally viewed from a substantially different angle are removed. The SIFT descriptors of the remaining points are then matched to those of the current image features to obtain initial 2D-3D correspondences. The 3D pose of the current frame is then estimated using Perspective-n-Point (PnP) localization \cite{lepetit2009epnp} embedded in a RANSAC outer loop.

PnP can fail if the video is discontinuous or if the viewpoint has changed drastically between frames. If the number of PnP inliers is too low, the submap is terminated at the previous frame and a new submap is initiated at the current frame.

\textbf{Keyframes.}
As new parts of the world come into view, the number of PnP inliers continually decreases. When this number falls below a threshold $\tau_{resection}$, the system adds a new keyframe and triangulates new 3D points. Selection of $\tau_{resection}$ is key in ensuring a balance between speed and stability: a high value causes frequent keyframe additions and thus slows the overall system, while a low value reduces visual overlap between keyframes and may lead to unstable or failed bundle adjustment. 
%For the datasets presented in this paper, $\tau_{resection}$ is set at 500.

To improve stability further, an additional keyframe is chosen halfway between the previous keyframe and the current frame. This \textit{middle} keyframe's pose is initialized to its PnP estimate, and its 2D PnP inliers are added as observations of their counterpart 3D landmarks for BA.  Geometrically consistent 2D matches are obtained between the middle keyframe and the current frame via RANSAC-based essential matrix estimation, and these matches are triangulated using the PnP pose estimates to form new 3D landmarks and corresponding 2D observations for BA. The SIFT descriptor from the \textit{middle} keyframe is assigned to each landmark.

Since the triangulation of new landmarks does not take into account the landmarks that have already been constructed, a subset may be duplicated. Therefore, any new landmark whose descriptor matches any existing landmark is removed. The current frame is finally added as a keyframe and its pose and observations are added to BA.

This selection strategy depends on neither physical nor temporal distance between keyframes; instead, keyframes are added only when sufficient new visual information is available, allowing the system to process videos acquired at arbitrary speed. In the Downtown dataset, for instance, the system selected keyframes spaced roughly 25 to 100 frames apart.

\begin{figure*}
\setlength{\tabcolsep}{1pt}
\begin{tabular}{llll}
%\fbox{\includegraphics[width=0.8\linewidth]{overview.png}}
\includegraphics[width=4.3cm]{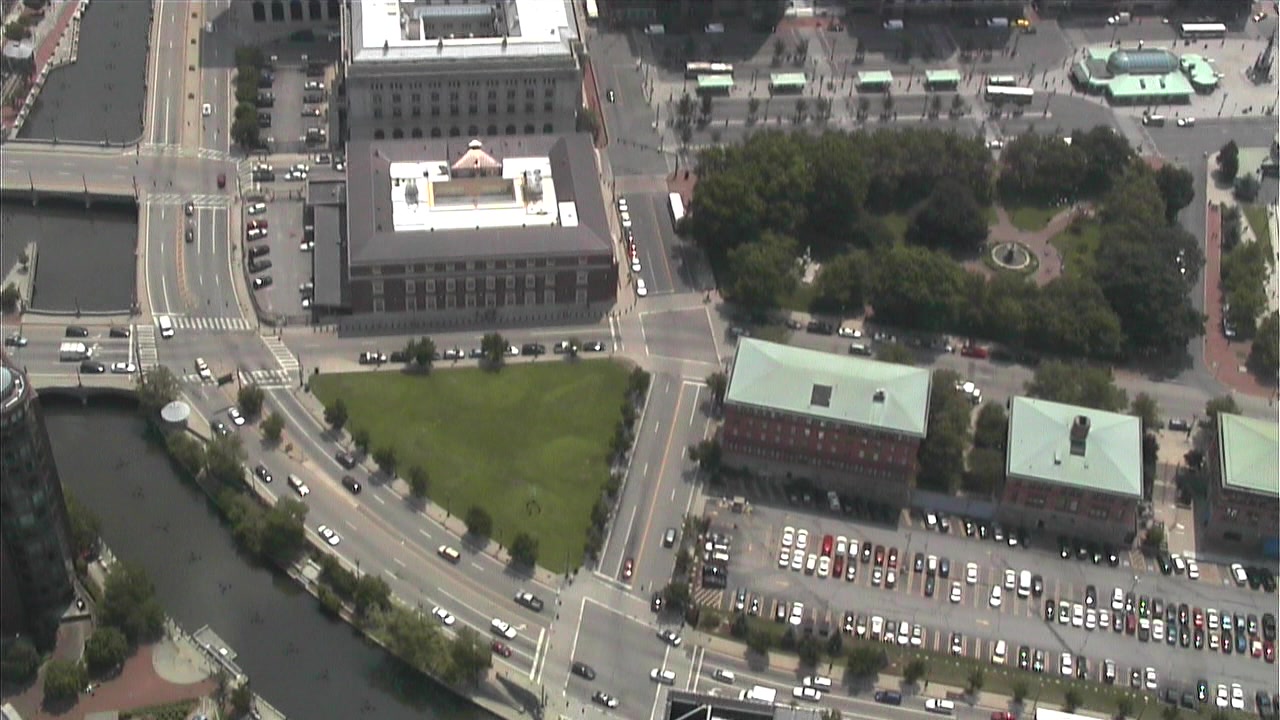}
\includegraphics[width=4.3cm]{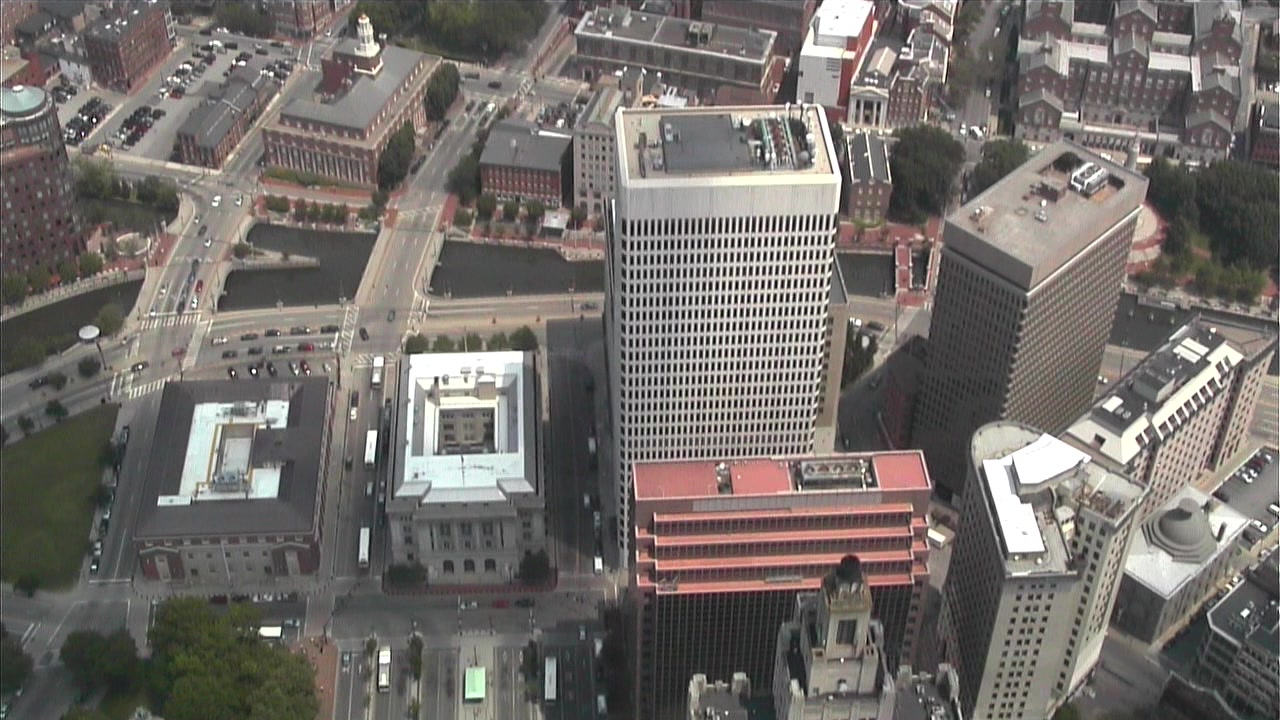}
\includegraphics[width=4.3cm]{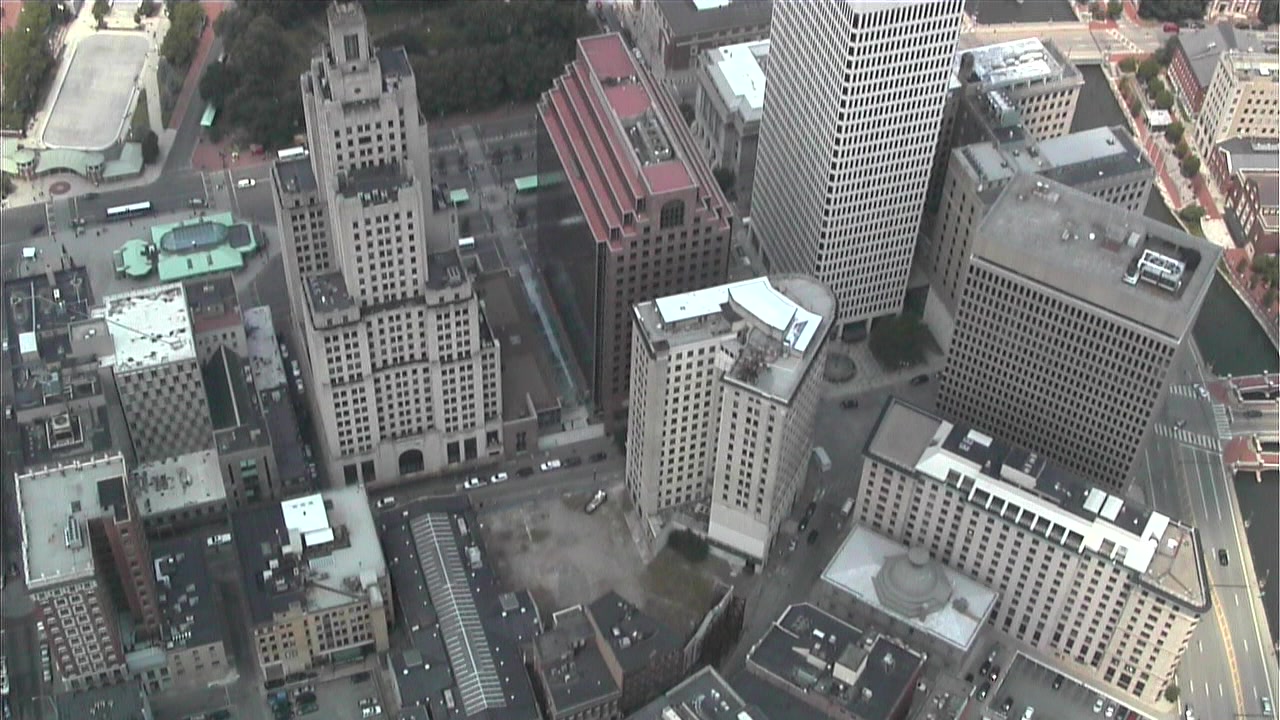}
\includegraphics[width=4.3cm]{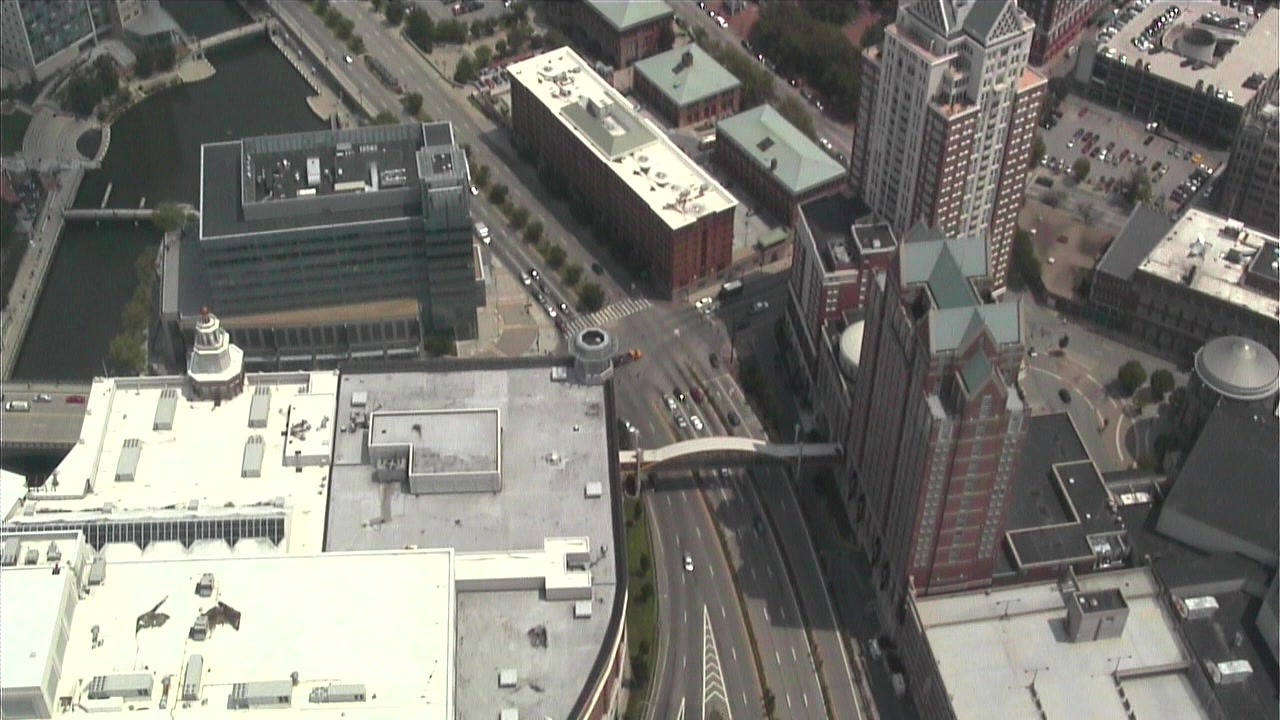}
\\
\includegraphics[width=4.3cm]{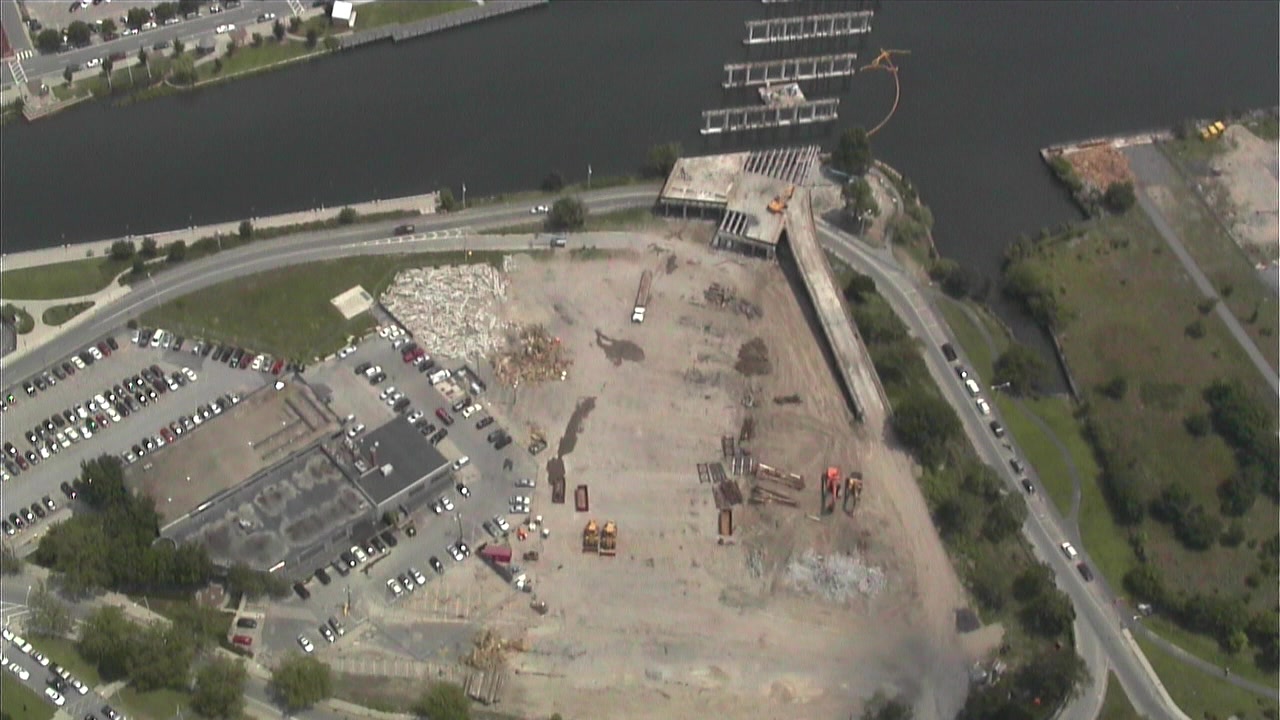}
\includegraphics[width=4.3cm]{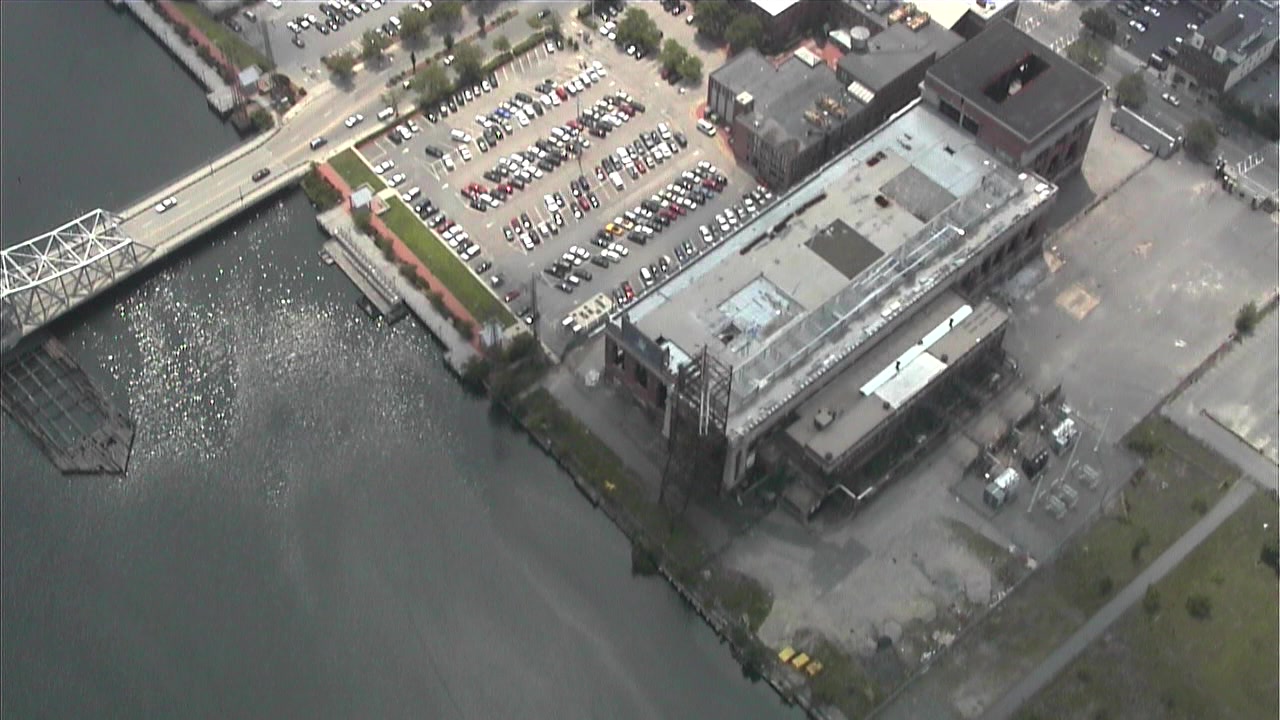}
\includegraphics[width=4.3cm]{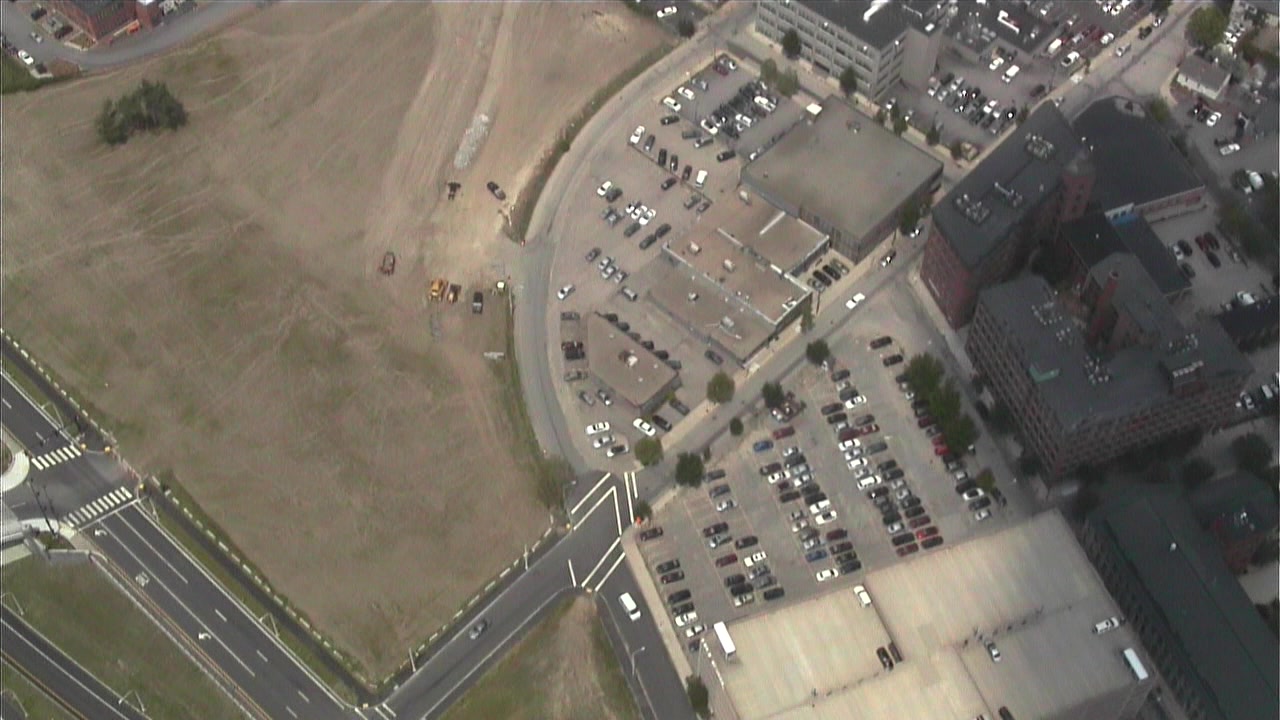}
\includegraphics[width=4.3cm]{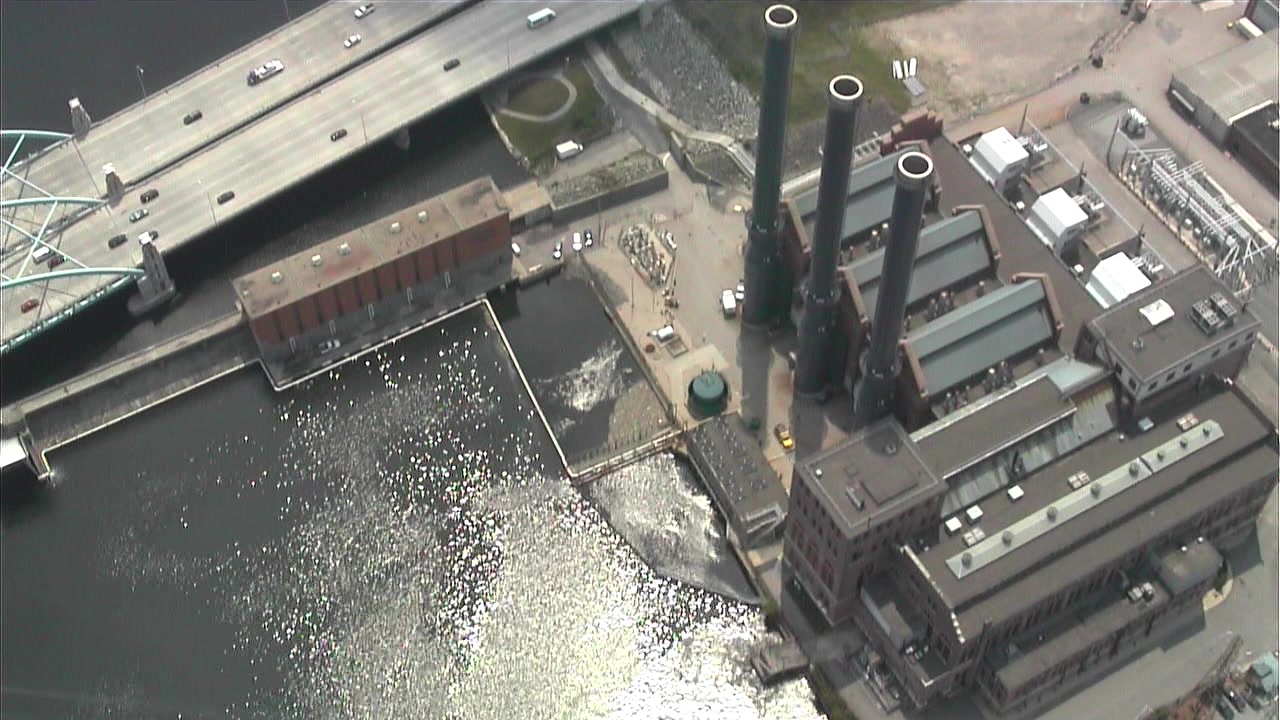}
\\
\end{tabular}
\caption{\textbf{Example images from the Downtown dataset.} A few scenes from the dataset showing challenging views with illumination changes, planar regions, high 3D relief, and water.}
\label{fig:frames}
\end{figure*}

\textbf{Incremental Bundle Adjustment.}
Similar to PTAM \cite{klein07parallel} and ORB-SLAM \cite{mur2015orb}, GLAM runs BA every time a new keyframe is added, optimizing the poses of all keyframes and the positions of all landmarks while holding fixed the intrinsic camera parameters. Different libraries including g2o \cite{kuemmerle11icra} and ceres-solver \cite{agarwal2013ceres} were evaluated, but pba \cite{wu2011multicore} performed best due to its specialization for BA problems.

\textbf{Bootstrapping}. When a new submap is initiated, no keyframes or 3D landmarks yet exist, so a strategy different from the steady-state process described above must be employed to establish initial geometry. The first frame of the submap becomes the first keyframe, with its pose fixed at the origin. Features from subsequent frames are matched against those of the first keyframe to find epipolar geometry via RANSAC-based essential matrix estimation \cite{Hartley00}. The current frame becomes the second keyframe if the number of RANSAC inliers falls below $\tau_{stereo}$ or if the average triangulation angle exceeds $\alpha_{stereo}$. (for the experiments described in this paper, we used $\tau_{stereo} = 1000$ and $\alpha_{stereo} = 30^{\circ}$). The inlier correspondences are triangulated to form the initial 3D landmarks.

\textbf{Completion.} A submap is sucessfully completed when the number of keyframes exceeds a threshold. We found that 20 keyframes were generally sufficient to form a stable reconstruction and minimize internal loop closures. Upon completion, all frames are re-localized to the final landmark set via PnP, which requires only that 3D-2D matches and not features to be recomputed. The next submap is initiated to have a degree of overlap (in our experiments, 10\%) with the current submap so that the two share a set of 3D points for alignment.

\textbf{Outlier filtering.}
Incorrectly matched point features lead to outliers that can dramatically affect bundle adjustment accuracy. Our system therefore filters outliers at several stages. First, after every bundle adjustment step, any point whose reprojection error exceeds a threshold is removed. Some incorrect matches may exhibit small reprojection error, such as when triangulation rays are nearly parallel and the reconstructed landmark is very far from the rest of the scene. To remove these points, the \textit{k} (30) nearest neighbours of each point are computed using a kd-tree, and a point is removed if its average distance from its neighbours is more than 2 standard deviations. Finally, upon submap completion, any point with large reprojection error in all frames is removed.

\begin{figure}[t]
\begin{tabular}{cc}
\includegraphics[width=8.2cm]{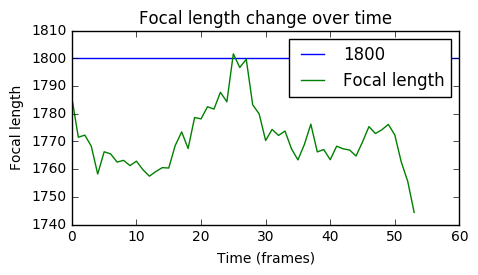}\\
(a)\\
\includegraphics[width=8.2cm]{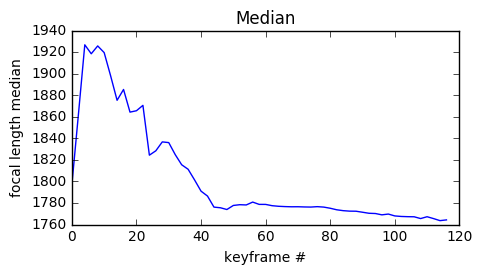}\\
(b)\\
\end{tabular}
\caption{\textbf{Estimation of focal length and radial distortion.} (a) Focal length of a new keyframe after it is added to bundle adjustment. (b) Median focal length of all keyframes after adding a new keyframe to bundle adjustment.}
\label{fig:focal}
\end{figure}

\textbf{Focal length estimation.}
Use of accurate camera intrinsics is instrumental in full-scale reconstruction, since there is no single rectifying transformation for incorrect intrinsics. The published focal length of 1800 for the lens used to collect the Downtown dataset was imprecise and led to reconstruction failures. We therefore included per-keyframe focal length in PBA optimization, observing that the median over all frames' estimated focal length converged to 1751 as the number of frames increased, as shown in Figure \ref{fig:focal}.

\subsection{Global Submap Alignment}
As local submaps are created, a parallel thread uses them as the unit of processing for full map reconstruction. This thread discovers correspondences between submaps, including loop closures, and optimizes pose and structure globally and efficiently.

\textbf{Submap matching.}
Although temporally adjacent submaps are known to be associated with one another by construction, it is required to discover visual links between submaps that view common scene content, regardless of the time at which they were created. This forms a connectivity graph in which submaps constitute nodes and commonly observed points constitute edges.

A naive implementation would identify visual links through brute-force search over every possible pair of submaps. This requires $O(n^2)$ time for $n$ submaps, making the process computationally intractable as $n$ grows. To form a connectivity graph efficiently, we employ a bag-of-words technique. A vocabulary tree is created offline using randomly-sampled SIFT descriptors drawn from aerial images; a series of k-means clustering operations partition the descriptors into a tree of visual words \cite{nister2006scalable}. As each submap is completed, its landmarks' SIFT descriptors are added to a database formed over the tree, and visual words are incrementally associated with the submap via an inverted index. The set of landmark descriptors in this submap is then used to query the database, which returns a ranked set of matching submaps and their weighted match scores. This mechanism reduces the complexity of submap matching from quadratic to linear time, as shown in Figure \ref{fig:realanalysis}.

The result is a graph whose edges represent potential visual matches based on an unordered bag of indexed visual words with no geometric constraints. Some graph edges may be incorrect, sharing a number of similar-looking features but not actually viewing common areas of the scene (e.g.\, due to repeated urban structures such as windows). The geometric consistency of correspondences is verified by estimating a closed form similarity transform between 3D landmarks in a RANSAC loop, accounting for both pose and scale differences. Edges with too few inliers are removed, as are all outlier landmark correspondences.

\begin{figure}
\begin{center}
%\bmvaHangBox{\includegraphics[width=5cm]{submap-graph2.png}}
{\includegraphics[width=6.0cm]{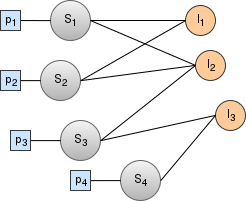}}
\end{center}
   \caption{\textbf{Representation of a submap pose-graph.} The nodes $S_1$, $S_2$, $S_3$,... represent a 7 DoF Sim(3) transformation for each submap. The nodes $l_1$, $l_2$, $l_3$,... represent a location in $\mathbb{R}^3$ for each 3D landmark point. An edge between a submap and 3D point represents the observation of the point in the submap. The nodes $p_1$, $p_2$, $p_3$,... are scale prior nodes for every submap that prevent collapse to zero scale during optimization. }
\label{fig:submapgraph}
\end{figure}

\textbf{Submap pose-graph}
After establishing temporal and spatial correspondences between submaps, the next step is to create a pose-graph and optimize the pose of the submaps such that the 3D distance between corresponding points is minimized (Figure \ref{fig:submapgraph}). Scale-drift \cite{strasdat2010scale} within submaps is addressed implicitly by the keyframe addition strategy, while scale-drift across submaps is adressed by using a 7 DoF similarity transform to represent each submap's pose.

The distance between corresponding 3D points of the submaps is minimized by using non-linear Gauss-Newton (GN) optimization \cite{Hartley00}, which involves finding a Jacobian w.r.t. to all free parameters. In this case, the 7 DoF pose of a submap and the $\mathbb{R}^3$ location of a landmark are the free parameters of the cost function. Differentiating the cost function w.r.t. the Lie group elements \textit{(R,s,t)} yields the Lie algebra space sim(3) \cite{eade2014lie}. sim(3) is a 7 dimensional vector space ($\omega$, $\sigma$, $\mu$), where $\omega$, $\sigma$ and $\mu$ respresent the Lie algebra components of rotation, translation and scale respectively.

Given a set of submaps $S$ and landmarks $L$, a 7 DoF similarity transform $(R, s, t)$ is associated with every submap $i$ and represented in the pose-graph with a Sim(3) node. A world point $X_j$ is represented with a $\mathbb{R}^3$ node in the pose-graph. If submap $i$ observes 3D point $X_j$, it is represented by edge $x_{ij}$ in the pose-graph.
Scale is constrained by multiplication of a small constant $\lambda_{a}$ ($<$1) with the Lie algebra of scale, $\mu_i$.
The scale prior prevents the solver from converging to a trivial solution, i.e zero scale.
The resulting cost function is:

\begin{equation}
	f(S,L) = \sum_{i \in S} \sum_{j \in L} ||s_i R_i x_{ij} + t_i - X_j||^2 + \sum_{i \in S}|| \lambda_{a} \mu_{i} ||^2
\end{equation}

%A 7 DoF Sim(3) matrix is over-parametrized using a 4x4 transform, which is a Lie group representation. But its update at every step of optimization is a 7-vector $\upsilon$ ($\omega$, $\sigma$, $\mu$) in the Lie algebra space.
%\\ An often used scale prior subtracts the scale from 1.\\ 
%TODO: Fix this
Initially, the scale prior used was $\sum_{i \in S} ||c - s_{i}||^2$, where c was a small constant, typically 1. When added to the remaining cost function, this scale prior prevented the trivial solution by evaluating to a non-zero value. However, it is an invalid operation since it is subtracting the Lie group $s_{i}$ from 1, and addition / subtraction operations are undefined for Lie groups. Hence, a valid scale prior instead constraints the scale by multiplying the Lie Algebra of scale $\mu$ with a small constant $\lambda$. This scale prior has a similar effect on the overall cost function at $\mu$ equals zero. It was also observed experimentally that a cost function with the former scale prior resulted in submaps that didn't align properly to each other but the latter scale prior fixed those alignment issues.

Every step of a GN optimization updates the sim(3) vector of each submap vertex with a $\upsilon_{up}$ vector. If $\upsilon$ and $G$ are the Lie algebra and Lie group of the current state of a Sim(3) vertex, then the vector update $\upsilon_{up}$ is applied as follows:
\begin{align}
G_{up} &= exp(\upsilon_{up}) \\
\upsilon &= log(G_{up} * exp(\upsilon)) \\
G &= G_{up} * G
\end{align}

The g2o \cite{kuemmerle11icra} framework is used to form and optimize the factor graph via Levenberg-Marquardt (LM), with robustness added via Huber loss \cite{Hartley00}.  Preconditioned conjugate gradient is used to efficiently solve the sparse linear system of equations arising at each step of LM, and analytic Jacobians were dervied to further speed up the computation. and are as follows:

\begin{align}
\frac{\partial{y}}{\partial{R}} = -y_{\times},   \hspace{2mm} 
\frac{\partial{y}}{\partial{t}} = I,    \hspace{2mm}
\frac{\partial{y}}{\partial{s}} = y
\end{align}
where, 
\begin{align}
y = (y_1, y_2, y_3) = f(S,L)
\\
y_{\times} = 
\begin{bmatrix}
0 & -y_3 & y_2 \\
y_3 & 0 & -y_1 \\
-y_2 & y1 & 0
\end{bmatrix}
\end{align}

\begin{figure*}
\begin{center}
\begin{tabular}{cc}
\includegraphics[width=.5\textwidth]{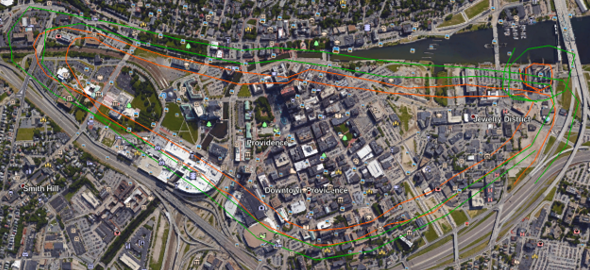}&
\includegraphics[width=.5\textwidth]{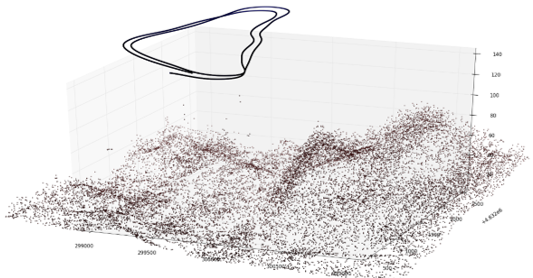}\\(a) & (b)\\
\includegraphics[width=.4\textwidth]{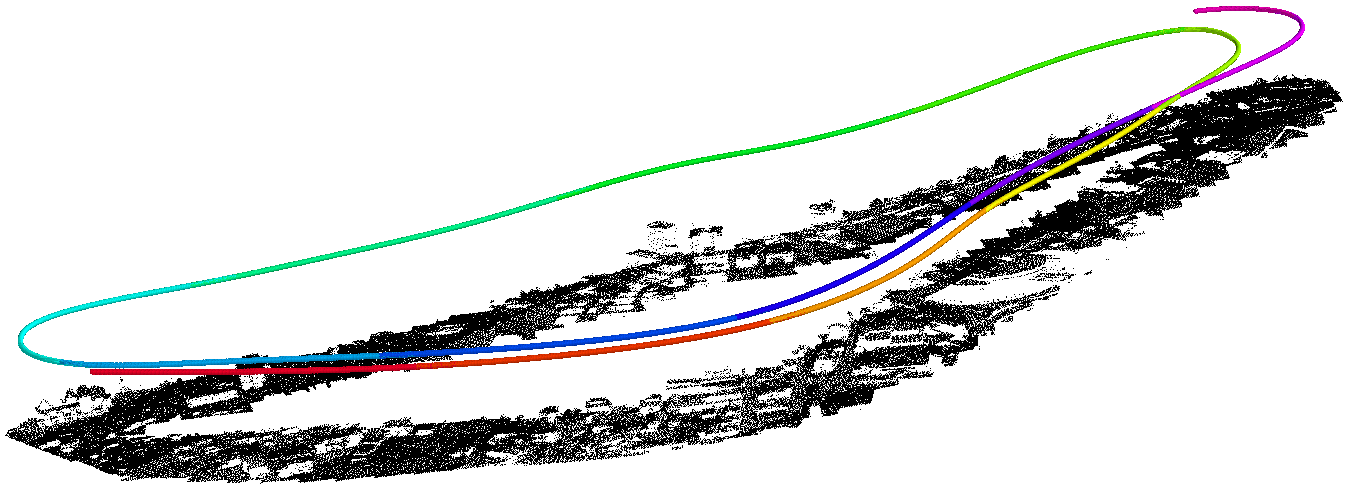}&
\includegraphics[width=.4\textwidth]{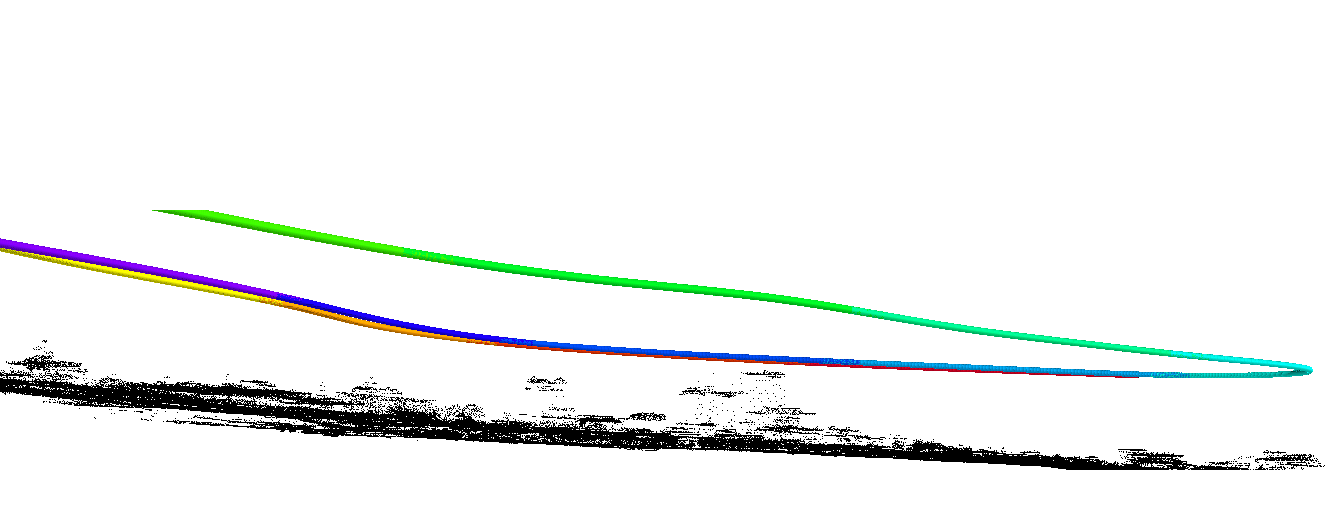}
\\(c) & (d)\\
\end{tabular}
\end{center}
\caption{\textbf{Synthetic datasets.} (a) Two camera trajectories and viewing direction trajectories drawn in green and red respectively in Google Earth. (b) LIDAR points and the synthetic camera and viewing direction trajectories of synthetic dataset synth1. (c), (d) 3D reconstruction of synth1 using GLAM. The cameras from each submap are represented with a different colour. Results from datasets synth2 and synth3 are shown in the supplementary material.}
\label{fig:synthfig}
\end{figure*}

\section{Experiments}
We evaluated GLAM on both synthetic and real datasets. Our experiments demonstrate centimeter level accuracy on synthetic datasets, and accuracy comparable to VisualSFM \cite{wu2011visualsfm} on real datasets, while providing a much faster run-time. 

\subsection{Synthetic Data}
We created synthetic datasets using Google Earth to simulate camera trajectories and using a LIDAR point cloud \cite{rigis} to serve as ground truth for simulated image projections. The results of 3D reconstruction on three such datasets is shown in Figure \ref{fig:synthfig} and summarized in Table \ref{tab:synthetic}.

\begin{table}[!htb]
\begin{center}
\begin{tabular}{|l|c|c|c|c| }
\hline
 & Frames & Submaps & Points & RMSE(cm)\\
\hline
synth1 & 1100 & 14 & 239727 & 0.3 \\ \hline
synth2 & 3408 & 15 & 328337 & 3.4 \\ \hline
synth3 & 40000 & 53 & 505000 & 0.6  \\
\hline
\end{tabular}
\end{center}
\caption{\textbf{Results on synthetic datasets.} Comparison of reconstructed maps to ground truth demonstrates errors on the order of centimeters and millimeters. The number of points in synth3 was made intentionally smaller than those in synth1 and synth2. Figures of the reconstruction from the three synthetic datasets can be found in the supplmentary material.} 
\label{tab:synthetic}
\end{table}

\begin{figure*}
\setlength{\tabcolsep}{.16667em}
\begin{tabular}{ccc}
\includegraphics[width=5.7cm]{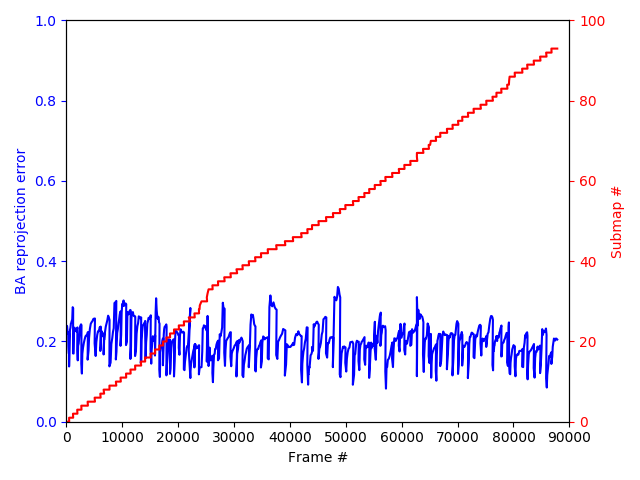}&
\includegraphics[width=5.7cm]{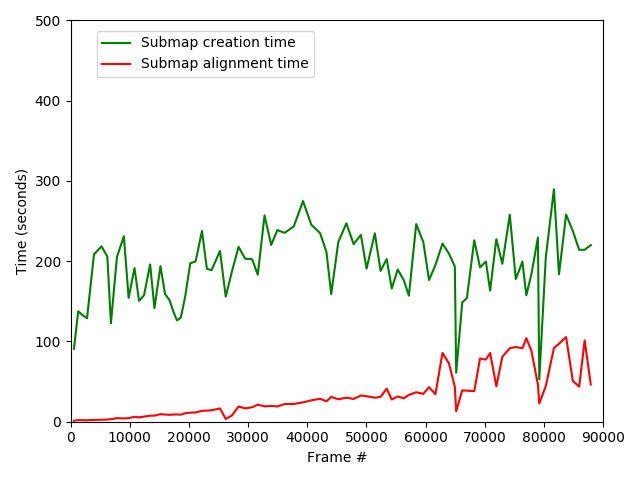}&
\includegraphics[width=5.7cm]{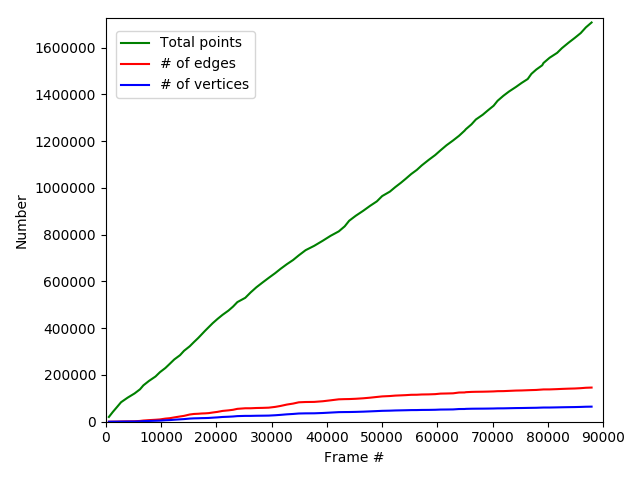}\\
(a)&(b)&(c)
\end{tabular}
\caption{\textbf{Analysis of the system on the Downtown dataset.} (a) The final bundle adjustment reprojection error within each submap (blue) remains bounded as new submaps are created every few thousand frames (red). (b) Submap creation time (green) also remains bounded, while submap alignment time (red) increases linearly but maintains acceptable latency for real-time operation. Spikes in the red curve are due to convergence failure when the graph has multiple disconnected components that are resolved later when missing correspondences are established. (c) At 88,000 frames, the total number of 3D points in the map are 1.7 million but the number of vertices and edges in the submap pose-graph are just 64k and 146k respectively, which allows an optimization of the full graph in real-time. A naive method that does incremental bundle adjustment with g2o on all the points in the dataset would have to potentially deal with BA on 1.7 million points, which would make each step of bundle adjustment infeasible in real-time and the required memory would potentially not fit on a single consumer-grade machine.}
\label{fig:realanalysis}
\end{figure*}

\subsection{Real Data}
\begin{figure*}
\begin{tabular}{ccc}
\includegraphics[width=8.2cm]{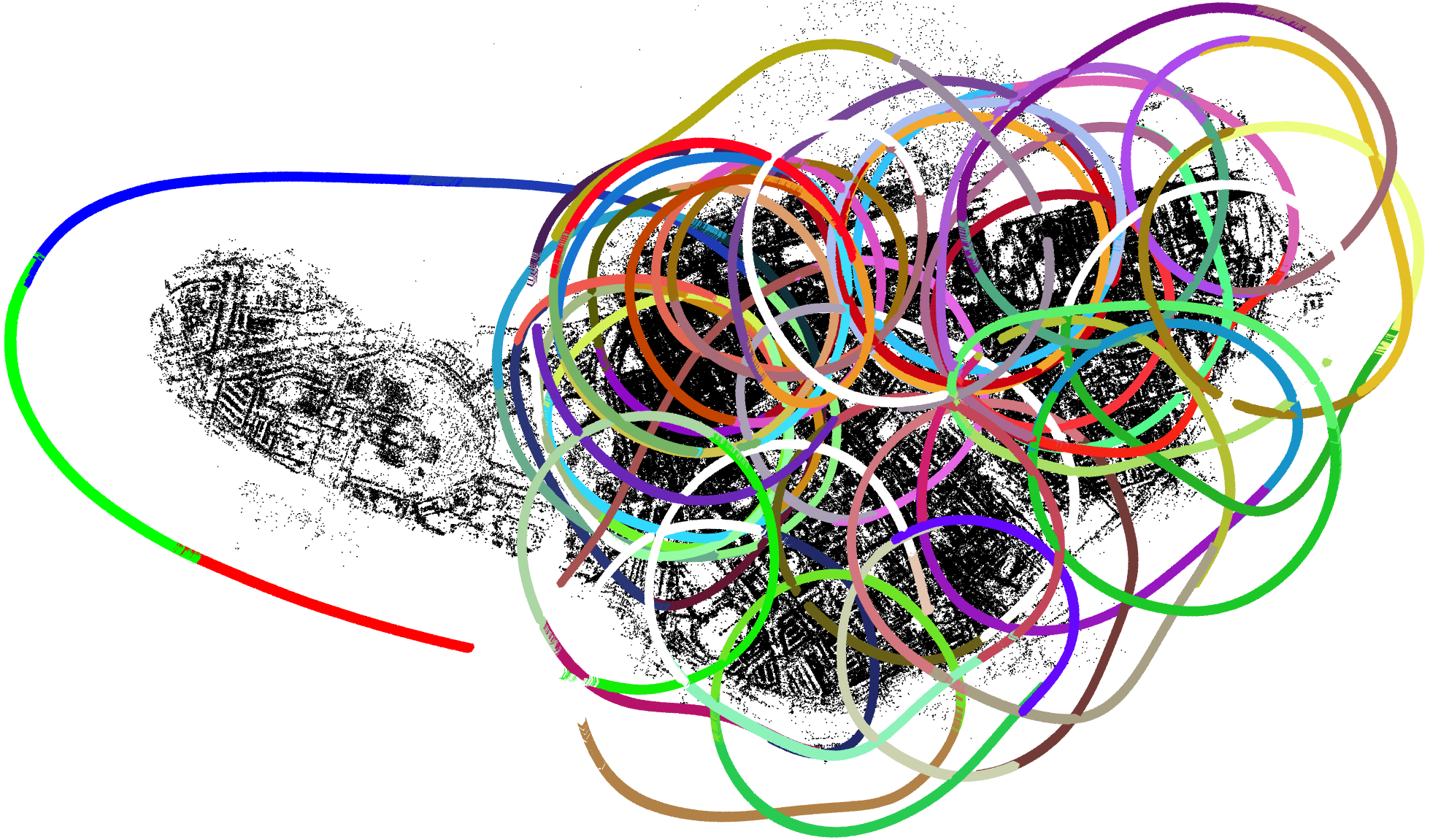}&
\includegraphics[width=8.2cm]{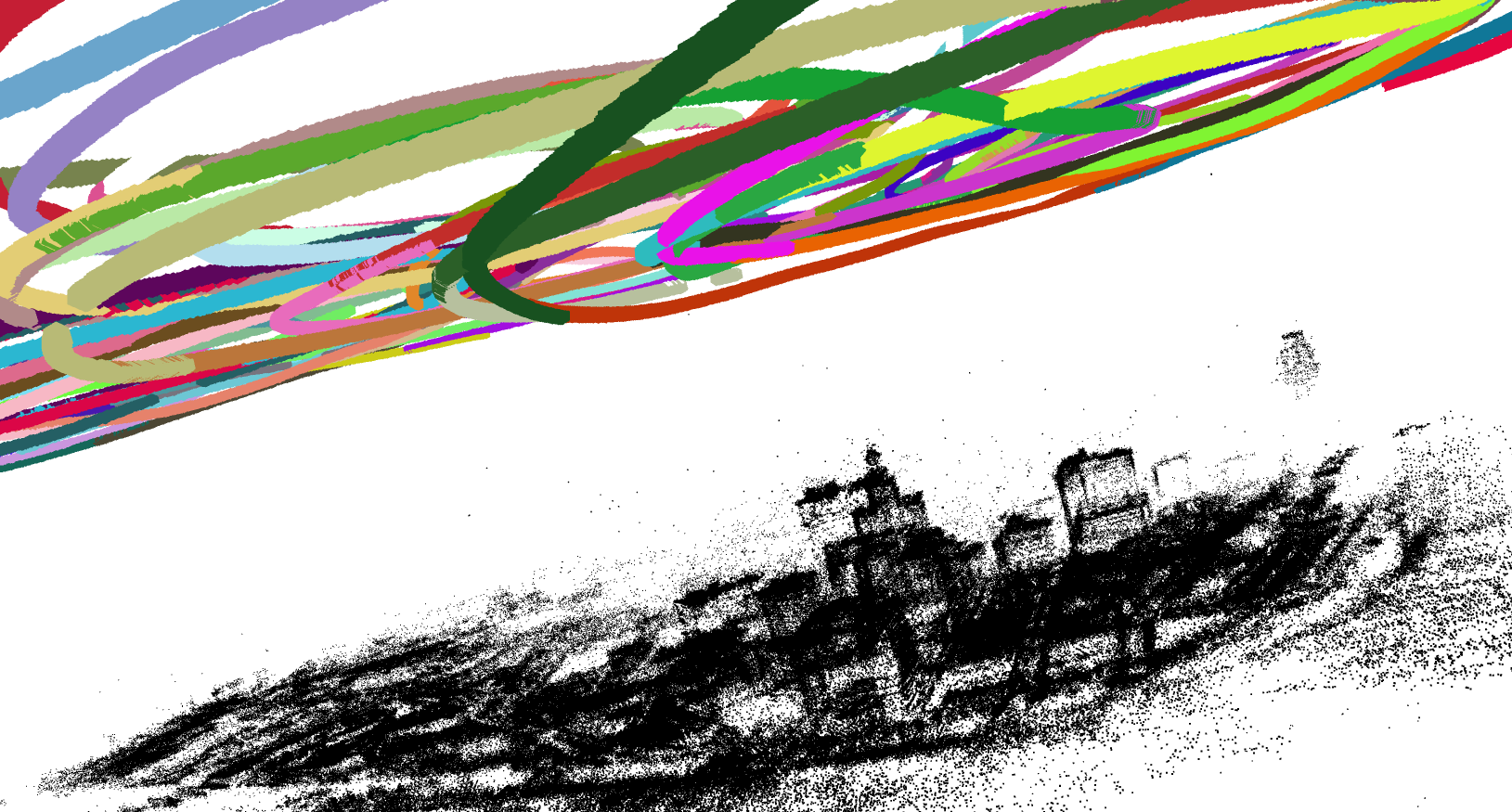}\\
(a)&(b)\\
\includegraphics[width=8.2cm, angle=0]{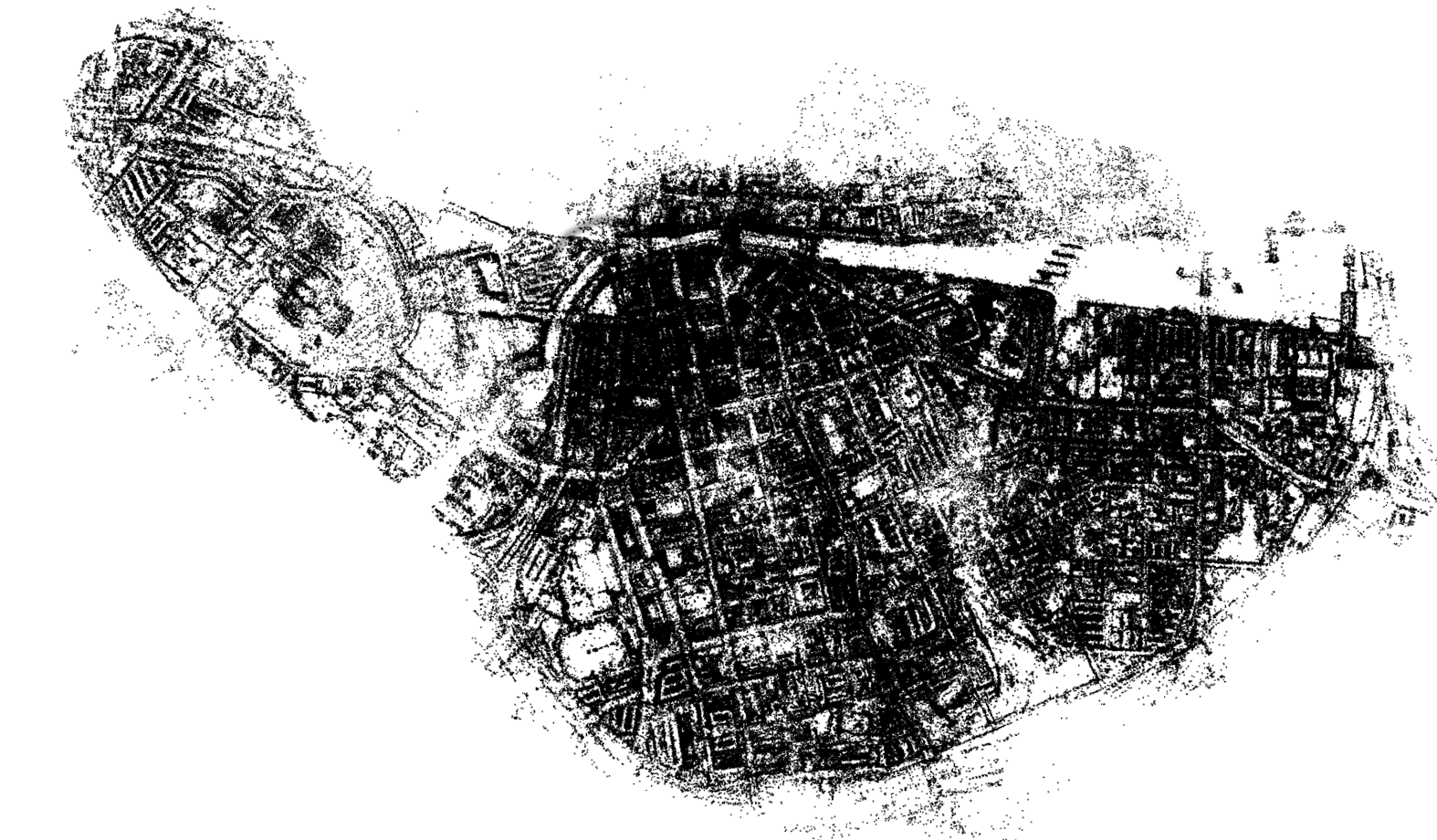}&
\includegraphics[width=8.2cm]{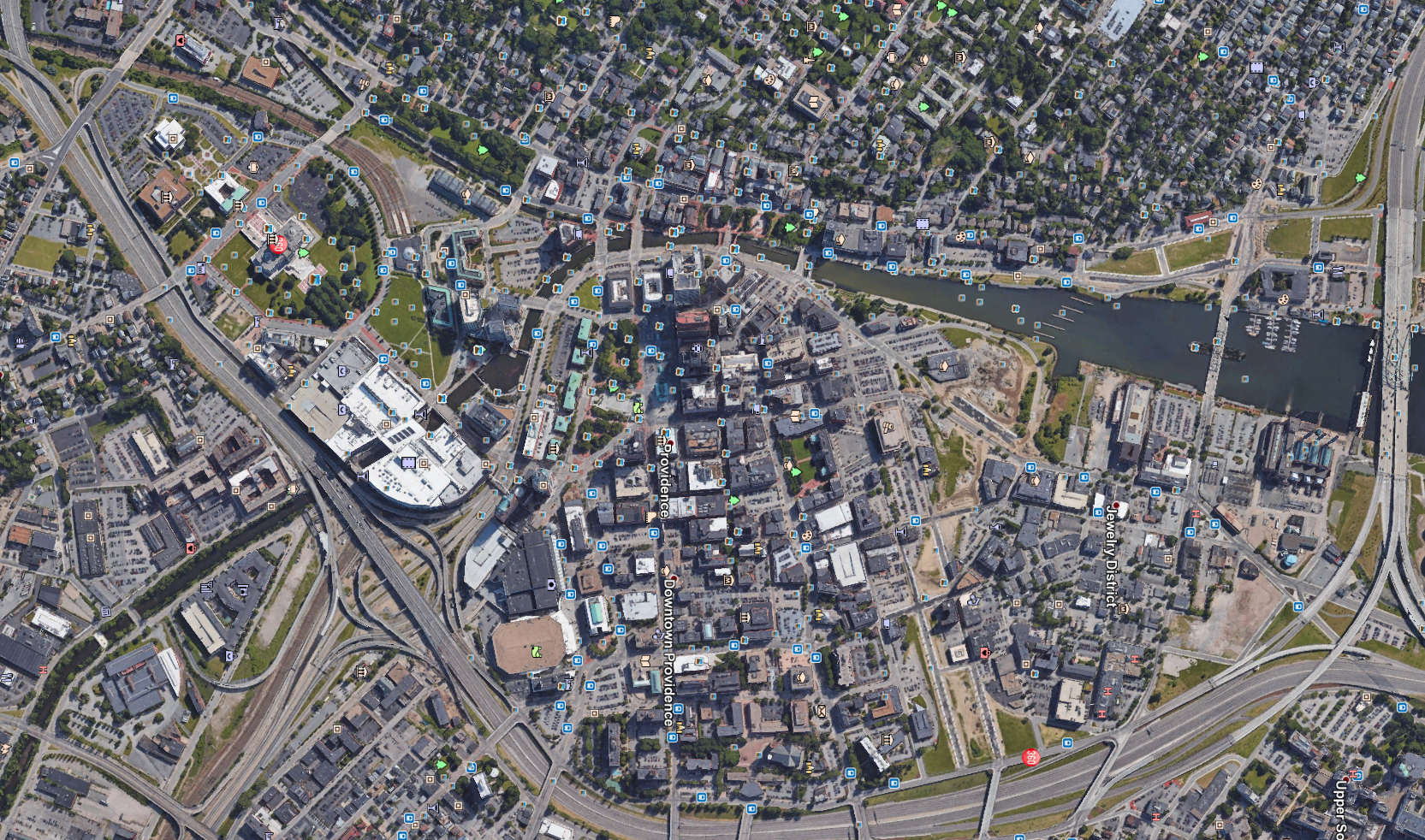}\\
(c)&(d)
\end{tabular}
\caption{ \textbf{Final results.} (a), (b) GLAM created 93 submaps for the full Downtown sequence, but 6 submaps failed in early stages due to bad frames. The above figure shows the 87 successful submaps aligned over the Downtown dataset. The cameras of each submap are represented with a different color. (c) The reconstructed 3D point cloud. (d) Google Earth image of Providence Downtown.}
\label{fig:realresults}
\end{figure*}

The Downtown dataset (Figure \ref{fig:frames}) is a video of 88,100 frames shot over Providence, USA. We applied GLAM to the entire video; Figure \ref{fig:realanalysis} illustrates runtime performance and scale, and Figure \ref{fig:realresults} visualizes the reconstruction results. The average run-time for submap creation is 5.8 fps, while the latency of the submap alignment and reconstruction thread is typically a few seconds.

We registered the landmark cloud reconstructed in the first 5,000 frames to a LIDAR point cloud of the same area using ICP \cite{besl1992method} to quantify accuracy with respect to ground truth. We also compared performance with VisualSFM \cite{wu2011visualsfm}, run on every $10^{th}$ frame of the first 5,000 frames (Table \ref{tab:realresults}). 

\begin{table}[!htb]
\begin{center}
\begin{tabular}{|l|c|c| }
\hline
& RMSE & Runtime\\
\hline
Visual SFM & \textbf{2.88 m} & 3.96 hrs \\ \hline
GLAM & 2.93 m & \textbf{11.79 min} \\ \hline
\end{tabular}
\label{tab:realresults}
\end{center}
\caption{\textbf{Comparison with Visual SFM.} ICP results on the first 5k frames of Downtown datset. GLAM is much faster than Visual SFM with comparable accuracy. Note that the LIDAR points are quantized to one meter spacing.}
\end{table}

\section{Conclusion}
This paper introduces a new visual SLAM system that can process long videos in near real-time on a single computer. Local submaps are constructed using incremental bundle adjustment over keyframes, and the global scene is reconstructed using factor graph optimization over submaps rather than keyframes, which allows closure of large loops. Submap creation has constant runtime complexity throughout, while submap alignment time increases linearly at a low rate, indicating that the system can be run on even larger datasets while maintaining acceptable latency.

The system's main performance bottleneck is currently SIFT feature extraction and matching. Future work will focus on improving run time by using binary descriptors. We also plan to investigate reducing the total number of constraints, incorporation of global averaging techniques, testing the approach on even larger datasets, and applying the system to terrestrial videos.

\section{Acknowledgements}
The authors would like to thank Vishal Jain his helpful advice and guidance on different parts of the project. This research is based upon work partly  supported  by  Air  Force  Research  Laboratory(AFRL)  under  contract  number FA8650-14-C-1826. This document is approved for public release via 88ABW-2017-2724.

{\small
\bibliographystyle{ieee}
\bibliography{egbib}
}

\end{document}